\documentclass[11pt]{article}

\usepackage[preprint]{acl}

\usepackage{times}
\usepackage{latexsym}
\usepackage[T1]{fontenc}
\usepackage[utf8]{inputenc}
\usepackage{microtype}
\usepackage{inconsolata}

\usepackage{graphicx}
\usepackage{booktabs}
\usepackage{enumitem}
\usepackage{multirow}
\usepackage{makecell}
\usepackage{amssymb}
\usepackage{algorithm}
\usepackage{algpseudocode}
\usepackage{amsmath}
\usepackage{float}

\usepackage{hyperref}

\title{EFlow: Learning Evidence Flow for Long-Video Reasoning\\with Adaptive Reflection}
\author{
  \textbf{Wenhao Zhang\textsuperscript{1,*}} \quad
  \textbf{Kuanwei Lin\textsuperscript{1,*}} \quad
  \textbf{Xuyi Yang\textsuperscript{2}} \quad
  \textbf{Wei Gao\textsuperscript{1}} \quad
  \textbf{Ge Li\textsuperscript{1,\textdagger}}
  \\
  \textsuperscript{1}School of Electronic and Computer Engineering, Peking University
  \\
  \textsuperscript{2}The Hong Kong University of Science and Technology
  \\
  \small{\texttt{\{whzhang25,gwlin25\}@stu.pku.edu.cn}}
}

\begin{document}
\maketitle
\makeatletter
\ifacl@anonymize\else
  \begingroup
  \renewcommand{\thefootnote}{\fnsymbol{footnote}}
  \footnotetext[1]{Equal contribution.}
  \footnotetext[2]{Corresponding author.}
  \endgroup
\fi
\makeatother

\begin{abstract}
Long-video reasoning is fundamentally constrained by how models acquire and utilize visual evidence. Existing tool-augmented video frameworks often interleave temporal grounding and answer reasoning within a single trajectory, causing early semantic hypotheses to bias evidence localization. We term this failure mode \textit{premature semantic commitment}, where biased grounding retrieves incomplete evidence and incomplete evidence further reinforces incorrect reasoning. To address this issue, we propose \textbf{EFlow}, an evidence-first video reasoning framework built upon Qwen3-VL. EFlow explicitly separates temporal grounding and logical reasoning through CoT for Temporal Grounding and CoT for Reasoning, enabling the model to retrieve relevant evidence before answer inference. In addition, EFlow introduces a confidence-aware reflection mechanism that re-evaluates the full video when retrieved evidence is potentially insufficient. We further construct dedicated trajectory datasets and train EFlow through supervised fine-tuning, reinforcement learning, and reinforcement fine-tuning. Extensive experiments across five video understanding benchmarks demonstrate that EFlow consistently improves long-video reasoning performance.
\end{abstract}

\section{Introduction}\label{sec:intro}

\begin{figure}[t]
\centering
\includegraphics[width=\columnwidth]{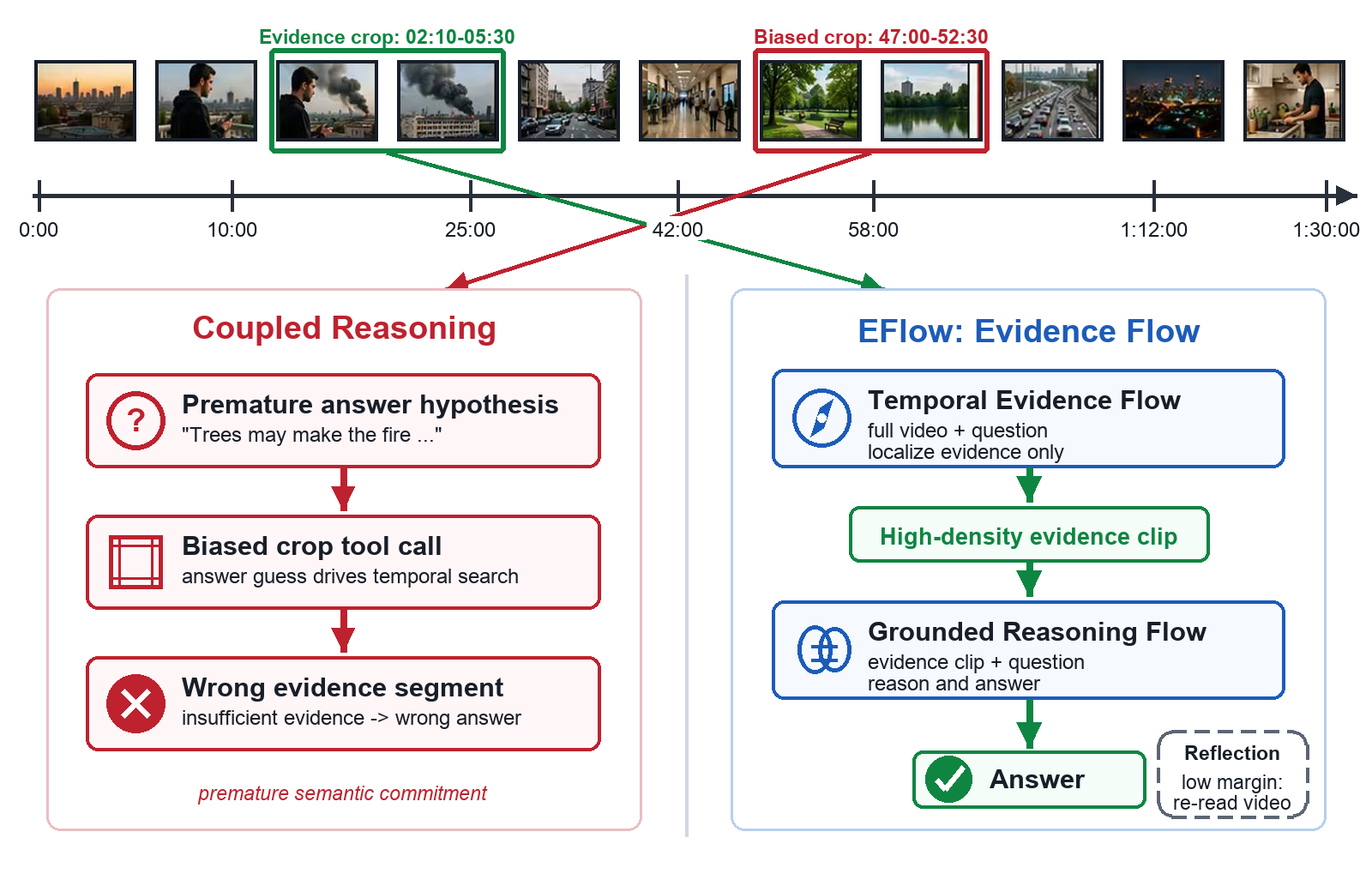}
\caption{Overview of \textbf{EFlow}. LongVT-style coupled reasoning can turn a premature answer hypothesis into a biased crop and a wrong answer. EFlow instead learns a transferable evidence flow: temporal grounding first localizes the evidence clip, grounded reasoning answers from the localized evidence, and adaptive reflection repairs low-confidence cases by re-reading the full video.}
\label{fig:overview}
\end{figure}

Long-video understanding requires more than recognizing visual content frame by frame. A model must follow a reliable process: deciding where to look, what evidence to preserve, when to reason, and when to distrust an initial conclusion. Although Large Multimodal Models (LMMs) have made rapid progress by aligning visual encoders with language models~\cite{qwen25vl, llavaonevision, llavamodel, videollava, videochatgpt, chatunivi, timechat}, the conventional single-turn paradigm---feeding uniformly sampled frames directly into the model---remains fragile as video duration grows. Redundant visual tokens dilute sparse evidence and strain the context window~\cite{jin2025videomem}, leading to hallucinations and loss of fine-grained temporal details~\cite{lvbench, longvideobench}. This suggests that long-video reasoning is not only a capacity problem, but also a process problem: the model must ground relevant evidence before committing to an answer.

Recent tool-augmented video frameworks address part of this challenge by using ``coarse-to-fine'' temporal tools, such as cropping or zooming, to navigate long videos before generating responses~\cite{videoagent, videotree, conan, longvt}. However, many such methods still interleave temporal grounding and answer reasoning within a single generation trajectory, allowing the model to form intermediate answer hypotheses before sufficient evidence has been retrieved. We identify this failure mode as \textit{premature semantic commitment}: early semantic assumptions bias temporal grounding, biased grounding retrieves incomplete or irrelevant evidence, and incomplete evidence further reinforces the initial wrong hypothesis. The resulting failure reflects the absence of a disciplined evidence flow that separates evidence acquisition from answer generation.

To address this issue, we propose \textbf{EFlow}, a training-and-inference framework for \textit{learning to flow}. Rather than introducing a new base model, EFlow teaches a model to acquire, transform, verify, and use video evidence through an explicit reasoning dynamic. During post-training, EFlow learns this transferable dynamic from structured trajectories and reward-guided optimization; during inference, it executes the learned dynamic as a staged evidence flow instead of producing a single entangled response.

At inference time, EFlow organizes this dynamic into a three-stage evidence flow. First, the \textbf{Temporal Evidence Flow} uses CoT for Temporal Grounding (T-CoT) to invoke a temporal cropping tool and localize visual evidence without predicting the final answer. Second, the \textbf{Grounded Reasoning Flow} uses CoT for Reasoning (R-CoT) to reason over the retrieved high-density evidence clip. These two learned stages align training and inference around the same principle: acquire evidence first and reason second, so that the model's reasoning behavior remains anchored to the visual evidence it has retrieved.

Beyond these learned flows, EFlow introduces a training-free \textbf{Adaptive Reflection Flow} as an engineered extension of inference. We compute a competition-based answer logit margin to estimate whether the retrieved evidence is sufficient. When the margin is low, EFlow treats uncertainty as a signal of potential evidence insufficiency and triggers a global ``re-reading'' step over the full video. This reflection path helps the model recover from insufficient evidence without requiring additional training.

We instantiate EFlow on Qwen3-VL~\cite{qwen3vl} with a three-stage post-training pipeline: Supervised Fine-Tuning (SFT) teaches the T-CoT/R-CoT trajectory format, Reinforcement Learning (RL) improves temporal grounding and answer correctness through outcome-based rewards, and Reinforcement Fine-Tuning (RFT) distills high-reward trajectories back into stable supervised behavior. Across five video understanding benchmarks, EFlow achieves strong performance, including 69.1\% on VideoMME Overall, 60.1\% on VideoMME Long, 52.5\% on LVBench, 60.3\% on LongVideoBench, and 80.0\% on NextGQA, demonstrating the value of explicit evidence flow for robust long-video reasoning.

Our contributions are summarized as follows:

\begin{enumerate}[leftmargin=2em]

\item We diagnose \textit{premature semantic commitment} as a central failure mode in tool-augmented long-video reasoning, and propose \textbf{EFlow}, a \textit{learning-to-flow} framework that separates evidence acquisition from answer reasoning.

\item We align training and inference through a staged flow design: T-CoT realizes the Temporal Evidence Flow, R-CoT realizes the Grounded Reasoning Flow, and dedicated trajectory data with SFT, RL, and RFT teach transferable evidence-flow dynamics.

\item We introduce a training-free confidence-aware reflection mechanism that uses answer logit margins to detect evidence insufficiency and adds a global re-reading path to the inference flow, yielding strong gains across diverse video understanding benchmarks.

\end{enumerate}

\section{Related Work}\label{sec:related}

\subsection{Thinking with Videos}\label{sec:related:thinking}

The concept of ``thinking'' in Large Multimodal Models (LMMs) has evolved from perception-only to multi-step reasoning with explicit intermediate thoughts. Building on textual Chain-of-Thought (CoT), recent works explore visual or multimodal CoT to guide step-by-step reasoning over images and videos. For video, models such as Video-R1~\cite{videor1}, Video-Thinker~\cite{wang2025video}, and VideoChat-R1~\cite{videochatr1} adopt R1-style reinforcement learning to encourage longer and more structured reasoning traces.

For long videos, however, plain CoT struggles: the model cannot retain all fine-grained details, and language-only reasoning easily drifts away from actual visual evidence. LongVT~\cite{longvt} addresses this by introducing an interleaved Multimodal Chain-of-Tool-Thought (iMCoTT) that alternates between global preview and temporal zoom-in via a native \texttt{crop\_video} tool, effectively ``thinking with long videos''. Most existing thinking paradigms, including these, still couple grounding and reasoning within the same reasoning trajectory, causing early semantic hypotheses to bias temporal grounding and relying on implicit self-correction. In contrast, EFlow organizes video reasoning as a staged evidence flow: T-CoT realizes Temporal Evidence Flow, R-CoT realizes Grounded Reasoning Flow, and adaptive reflection allows the model to re-read the video when retrieved evidence is potentially insufficient.

\subsection{Video Reasoning Frameworks}\label{sec:related:agents}

Tool-augmented frameworks extend LMMs with external tools such as web search or code interpreters~\cite{schick2024toolformer, shen2024hugginggpt, react, reflexion}. In the video domain, early systems like VideoAgent~\cite{videoagent} and VideoTree~\cite{videotree} repeatedly retrieve relevant clips before reasoning over the selected evidence. More recent frameworks, including Conan~\cite{conan}, Video-Zoomer~\cite{videozoomer}, VITAL~\cite{vital}, and LongVT~\cite{longvt}, integrate temporal tools such as \texttt{crop\_video} into model outputs, enabling multi-round global-to-local video inspection for long-video QA.

Most existing frameworks, however, still couple grounding and reasoning within the same reasoning trajectory, causing early semantic hypotheses to bias temporal grounding. In contrast, EFlow explicitly separates CoT for Temporal Grounding and CoT for Reasoning, and further introduces a confidence-aware reflection mechanism to re-evaluate the full video when retrieved evidence is insufficient.

\section{Methodology}\label{sec:method}

We propose EFlow, a video reasoning framework built on Qwen3-VL~\cite{qwen3vl} for long-form video question answering. EFlow combines temporal evidence localization, grounded answer generation, and confidence-aware verification within a unified inference procedure. Given a video of arbitrary length, it identifies a compact evidence segment with a native cropping tool, reasons over the localized visual evidence, and consults the full video context when the evidence is insufficient. This design is especially effective for long videos with sparse evidence and strained context windows.

\subsection{Architecture Overview}\label{sec:method:architecture}

\begin{figure*}[t]
\centering
\includegraphics[width=\textwidth]{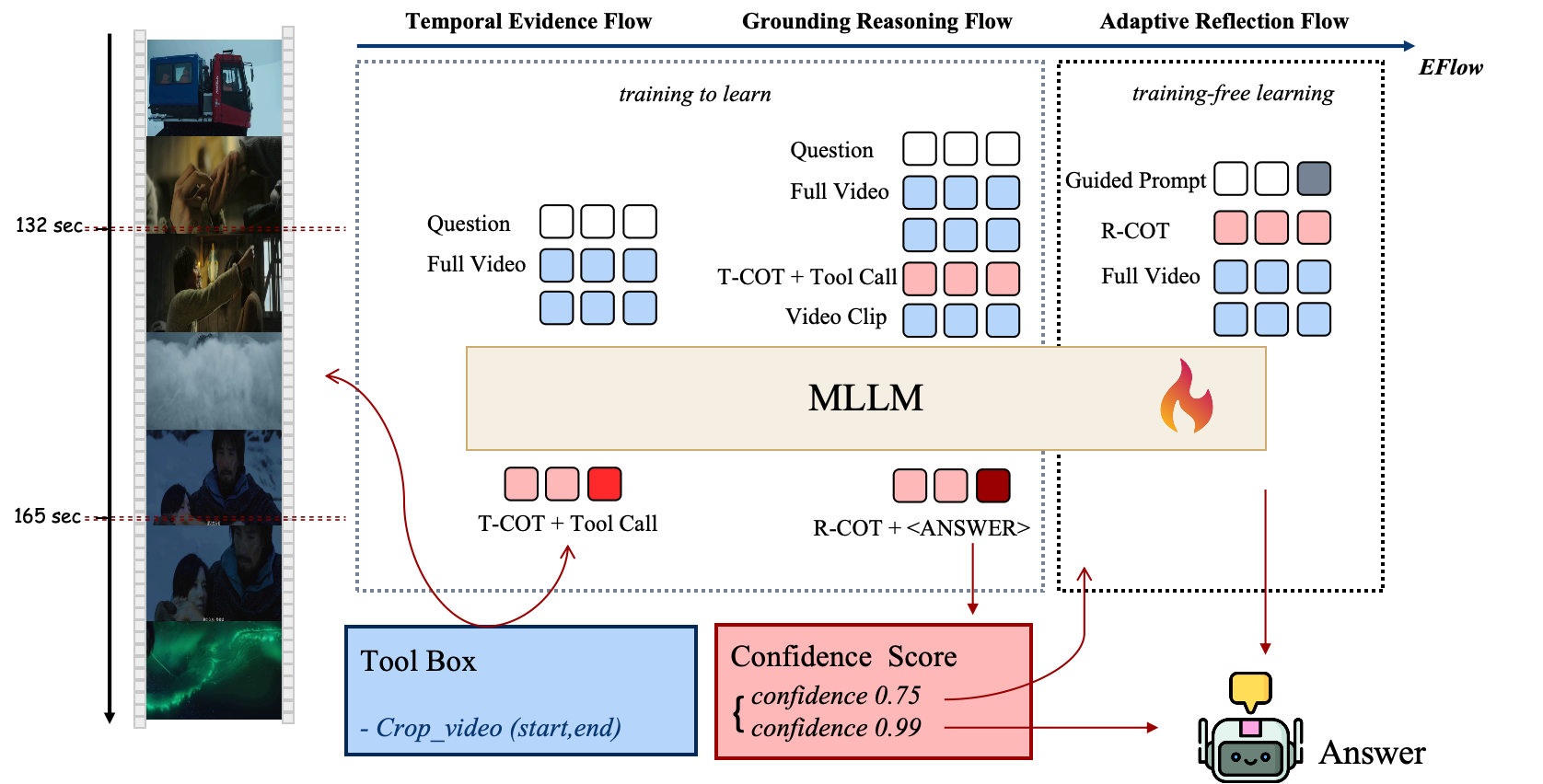}
\caption{Detailed architecture of EFlow. The framework organizes inference as an evidence flow: temporal grounding first inspects the full video and issues a crop command to obtain a candidate evidence clip; grounded reasoning then answers from the localized evidence and estimates confidence; and adaptive reflection re-reads the full video with the prior reasoning trace when the localized evidence appears insufficient. The T-CoT and R-CoT blocks in the figure denote intermediate trajectory formats used to supervise these stages, rather than separate model modules.}
\label{fig:method}
\end{figure*}

The core motivation of EFlow is to prevent premature semantic commitment during long-video reasoning by separating evidence acquisition from answer reasoning through specialized CoT sequences, followed by a confidence-driven reflection mechanism. As illustrated in Figure~\ref{fig:method}, the inference process is structured as three connected flows:

\textbf{Temporal Evidence Flow.} Given an input video $\mathcal{V} = \{v_t\}_{t=1}^T$ with $T$ frames and a user query $\mathcal{Q}$, the model processes the full video representation alongside the question. In this flow, the model is strictly constrained to focus on grounding and invokes a reasoning sequence we term CoT for Temporal Grounding (T-CoT). The LLM processes the scene and outputs a tool-calling command to extract a single highly relevant segment without attempting to output the final answer. This design explicitly prevents early semantic hypotheses from contaminating temporal grounding:
\begin{equation}
\begin{aligned}
(\text{T-CoT}, \mathcal{C}) 
&= \text{LLM}(\mathcal{V}_{full}, \mathcal{Q}) \\
\text{where} \quad 
\mathcal{C} 
&= \text{Crop\_video}(s, e)
\end{aligned}
\end{equation}
where $s$ and $e$ denote the start and end timestamps. The external toolbox then extracts the specific high-resolution clip: $\mathcal{V}_{clip} = \text{Crop\_video}(\mathcal{V}_{full}, s, e)$.

\textbf{Grounded Reasoning Flow.} The extracted clip provides a refined, high-density visual context. This evidence is fed back into the LLM alongside the original query, the full video, and the multi-modal trace from the Temporal Evidence Flow ($\text{T-CoT}$ and tool call). By separating evidence acquisition from answer reasoning, EFlow can reason over retrieved evidence without being biased by premature semantic assumptions. During this flow, EFlow generates a CoT for Reasoning (R-CoT) and the final answer $\mathcal{A}$. Crucially, we extract the logits at the exact decoding step where the answer label is generated. We calculate the softmax probabilities over all valid multiple-choice tokens (e.g., A, B, C, D) and define our confidence score $c$ as the probability margin between the top-1 prediction and the top-2 prediction:
\begin{equation}
    c = P(\tilde{y}_{top1}) - P(\tilde{y}_{top2})
\end{equation}
where $P(\tilde{y}) = \text{softmax}(\text{Logits}_{\text{answer\_step}})$. We denote $c$ as the \textit{margin} (probability difference), which serves as the Confidence Score. A low confidence margin often indicates that the retrieved evidence is insufficient or semantically ambiguous for reliable reasoning.

\textbf{Adaptive Reflection Flow.} We define an empirical confidence threshold $\tau$. 
To accommodate varying task difficulties, EFlow employs dynamic thresholding across benchmarks. 
If the confidence score $c \ge \tau$, the model is considered sufficiently confident and directly returns $\mathcal{A}$. 
Otherwise, when $c < \tau$, the narrow confidence margin suggests that the retrieved evidence may be insufficient due to premature semantic commitment during temporal grounding.

Rather than treating low confidence as a reasoning failure, we interpret it as a signal of potential evidence insufficiency. In this case, the Adaptive Reflection Flow re-injects the full video $\mathcal{V}_{full}$ together with the previous reasoning trace into the model for global re-evaluation. Specifically, let $\mathcal{R}$ denote the previous reasoning chain (R-CoT), and $\mathcal{P}$ denote the guided reflection prompt:
\begin{equation}
\mathcal{A}_{final}
=
\text{LLM}_{ref}
(\mathcal{V}_{full}, \mathcal{R}, \mathcal{P})
\end{equation}

This uncertainty-aware re-evaluation mechanism allows EFlow to recover from insufficient or biased evidence retrieval, thereby significantly enhancing the robustness of long-video reasoning. While our current confidence formulation is naturally suited for multiple-choice scenarios prevalent in standard video benchmarks, it can be readily adapted for open-ended reasoning tasks. For free-form generation, measuring the generation sequence's average probability or perplexity~\cite{liu2026videoauto} can seamlessly replace the discrete margin score to serve as the confidence heuristic for triggering the reflection phase.

\begin{algorithm}
\caption{Competition Confidence Computation}
\begin{algorithmic}[1]
\Require Output scores $\mathcal{G}$, sequences $\mathcal{S}$, tokenizer $\mathcal{T}$, options $\mathcal{L}$
\Ensure Prediction $p$, confidence score $c$, distribution $\mathcal{D}$

\State $t \leftarrow \text{LocateAnswer}(\mathcal{S}, \mathcal{T})$ \Comment{First $<$answer$>$X or post-$<$/think$>$}
\If{$t = \text{null}$} \Return $(\text{null}, 0, \{\})$ \EndIf

\State $\mathbf{z} \leftarrow \mathcal{G}[t][0]$ \Comment{Logits at answer step}

\For{$\ell \in \mathcal{L}$}
    \State $a \leftarrow \text{Extract}(\ell)$ \Comment{Answer letter}
    \State $\mathcal{V}_a \leftarrow \mathcal{T}(\{a, \text{~}a, >a\}).\text{ids}$ \Comment{Multi-variant tokens}
    \State $s_a \leftarrow \max_{j \in \mathcal{V}_a} \mathbf{z}[j]$ \Comment{Best matching logit}
\EndFor

\State $\mathbf{p} \leftarrow \text{Softmax}([s_a]_{a \in \mathcal{L}})$, sort descending: $p_{(1)} \geq p_{(2)}$
\State $c \leftarrow p_{(1)} - p_{(2)}$ \Comment{Margin-based confidence}
\State $p \leftarrow \arg\max_a p_a$, $\mathcal{D} \leftarrow \{a: p_a\}$

\Return $(p, c, \mathcal{D})$
\end{algorithmic}
\end{algorithm}

\subsection{Training Data Construction}\label{sec:method:data}
Training EFlow requires data that teaches not only the final answer, but also the intermediate evidence-flow behavior: where to look, how to invoke the crop tool, and how to reason from the localized evidence. We therefore construct three complementary datasets for Supervised Fine-Tuning (SFT), Reinforcement Learning (RL), and Reinforcement Fine-Tuning (RFT), each targeting a different stage of learning transferable evidence-flow dynamics.

\paragraph{EFlow-SFT-50K Dataset.}
We build our high-quality SFT dataset upon the foundation of the open-source Video-R1-CoT-165K datasets. However, standard CoT traces lack explicit tool-invocation logic. To inject the ``think with videos'' multi-turn capability, we design an automated generation pipeline with Gemini-3-Flash serving as a weak initialization model. First, we prompt Gemini-3-Flash to generate temporal crop boundaries and preliminary reasoning traces for each query. We strictly filter this initial pass, retaining only the samples where the predicted final answer matches the ground truth, followed by additional expert verification for temporal grounding consistency. Next, for these filtered samples, we apply a rule-based formatting prompt to rewrite the reasoning process into our T-CoT and R-CoT structural format (including the \texttt{<think>}, \texttt{<tool\_call>}, and \texttt{<answer>} tags). The generated trajectories further undergo human expert review to remove noisy samples and ensure the quality of evidence-grounded reasoning traces. This rigorous process yields a high-fidelity dataset of approximately 50,000 multi-turn trajectories, effectively teaching the base model how to stage temporal evidence acquisition before logical deduction.

\paragraph{EFlow-RL-10K Dataset.}For the RL phase, we curate a focused dataset of 10,000 queries primarily derived from the VideoITG-40K
\cite{wang2025videoitg} dataset, with the base videos sourced from the LLaVA-Video-178K~\cite{llavavideo} collection. Unlike the SFT stage, this dataset does not require full ground-truth tool-calling trajectories. Instead, by leveraging the inherent structure of VideoITG, each query provides the ground-truth final answer paired directly with a precise relevant temporal interval $[s_{gt}, e_{gt}]$. These intervals act as pure outcome-based supervision signals for the $r_{iou}$ and $r_{ans}$ rewards, allowing the GRPO algorithm to explore optimal, unconstrained grounding strategies autonomously.

\paragraph{EFlow-RFT-10K Dataset.}Finally, we construct a dataset for the reinforcement fine-tuning stage. We first filter the rollouts generated during the RL process based on their reward scores, selecting the top-ranked trajectories. These candidates then undergo expert review to ensure diversity in question types and video domains. The final collection is organized into 10,000 high-quality trajectory-level samples.

\begin{figure}[t]
\centering
\includegraphics[width=\columnwidth]{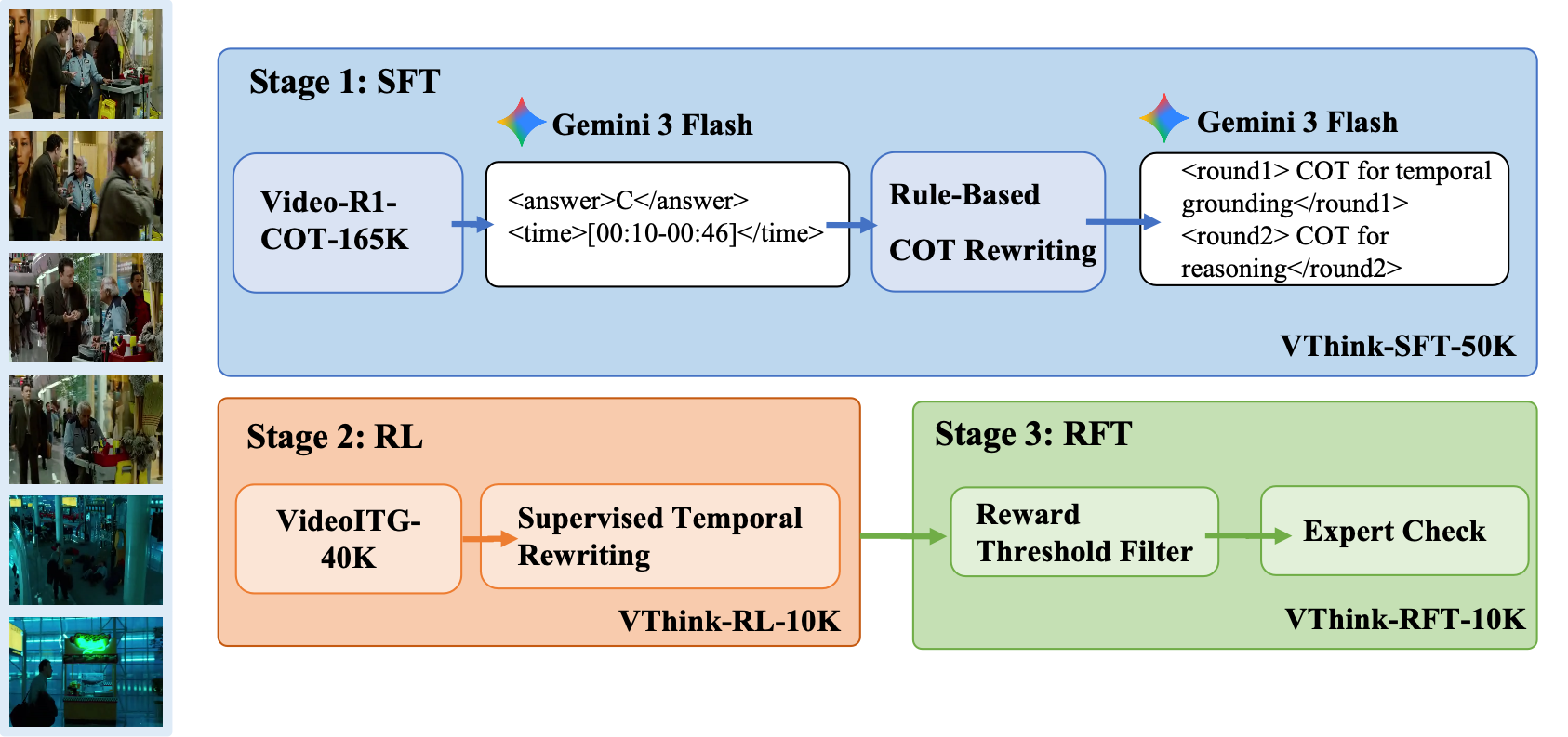}
\caption{Overview of the training data construction pipeline. We generate and filter Gemini-3-Flash temporal-boundary annotations to build EFlow-SFT-50K, and curate EFlow-RL-10K from VideoITG with ground-truth intervals for RL rewards.}
\label{fig:dataset}
\end{figure}

\subsection{Training Strategy}\label{sec:method:training}
We design a rigorous three-stage training pipeline comprising Supervised Fine-Tuning (SFT), Reinforcement Learning (RL), and Reinforcement Fine-Tuning (RFT).

\paragraph{Stage 1: Cold-Start Supervised Fine-Tuning.} In the first stage, the model is trained on the EFlow-SFT-50K dataset using the standard auto-regressive cross-entropy loss:
\begin{equation}
    \mathcal{L}_{SFT} = - \sum_{i} \log P(y_i | y_{<i}, \mathcal{V}, \mathcal{Q})
\end{equation}
where $y_i$ represents the tokens of the tool calls, reasoning steps, and the final answer. By fine-tuning Qwen3-VL on this diverse mixture, we initialize the model's ability to master the T-CoT and R-CoT structural generation, format tool calls correctly.

\paragraph{Stage 2: Agentic Reinforcement Learning.} While SFT teaches the model \textit{how} to use the tools, it does not necessarily teach it the \textit{optimal} strategy for complex, unseen videos. To unlock deeper exploration and self-correction without the prohibitive memory overhead of standard PPO, we employ Group Relative Policy Optimization (GRPO)~\cite{grpo} on the EFlow-RL-10K dataset. GRPO circumvents the need for a separate value model by sampling a group of trajectories for each query and optimizing the policy based on the relative advantages of the trajectories within the group.

To comprehensively guide EFlow, our rule-based total reward consists of three carefully designed components:

\noindent\textbf{Format Reward ($r_{format}$):} Continuously regulates the ``think with videos'' structural process. It assigns a base reward of $1.0$ if the model perfectly follows the multi-turn template by correctly opening and closing the \texttt{<think>}, \texttt{<tool\_call>}, and \texttt{<answer>} XML tags during both T-CoT and R-CoT phases. If the model generates malformed tags (e.g., missing closing tags or incorrect tool names), it receives a penalty of $0.0$.

\noindent\textbf{Temporal Grounding Reward ($r_{iou}$):} Evaluates the precision of the temporal grounding. Let $\mathcal{W}_{pred} = [s, e]$ be the model's predicted crop interval and $\mathcal{W}_{gt} = [s_{gt}, e_{gt}]$ be the ground-truth essential interval. We compute the temporal Intersection over Union (t-IoU), multiplied by a scaling factor $\lambda_{iou}$:
\begin{equation}
    r_{iou} = \lambda_{iou} \times \frac{|\mathcal{W}_{pred} \cap \mathcal{W}_{gt}|}{|\mathcal{W}_{pred} \cup \mathcal{W}_{gt}|}
\end{equation}
This continuous reward signal encourages the model to extract the tightest possible high-density clip containing the evidence, aggressively penalizing overly broad or entirely disjoint crops.

\noindent\textbf{Answer Accuracy Reward ($r_{ans}$):} Due to the complex, free-form nature of reasoning outputs, simple string matching is often overly strict and fails to capture semantically equivalent answers. Instead, we adopt an LLM-as-a-Judge approach. We employ DeepSeek-V3~\cite{deepseekv3} as an external critic to evaluate the semantic equivalence between EFlow's predicted answer $\mathcal{A}_{pred}$ and the ground-truth answer $\mathcal{A}_{gt}$. The judge assigns a discrete categorical score:
\begin{equation}
    r_{ans} =
    \begin{cases}
    1.0 & \text{if Full Complete Match;} \\
    0.5 & \text{if Partial Match / Ambiguous;} \\
    0.0 & \text{if Incorrect.}
    \end{cases}
\end{equation}

These fine-grained reward signals are summed as $R = r_{format} + r_{iou} + r_{ans}$ and fed into the GRPO framework, iteratively aligning the model's tool-use, temporal grounding, and final video reasoning behaviors without the need for dense token-level human annotations.

\paragraph{Stage 3: Agentic Reinforcement Fine-Tuning (RFT).} To further stabilize the agentic behaviors learned during RL and consolidate the model's reasoning capabilities, we introduce a third training stage: RFT. Following LongVT~\cite{longvt}, we use the term RFT to refer to supervised fine-tuning on high-reward trajectories produced by the reinforcement learning stage, rather than another round of online RL optimization. In this stage, we utilize the EFlow-RFT-10K dataset constructed via reward filtering and expert verification. We conduct supervised fine-tuning on these high-quality trajectories to effectively distill the successful exploration patterns back into the model parameters. This self-distillation process helps the model internalize robust grounding and tool-calling behaviors, pushing its performance beyond the ceiling of standard SFT.

\begin{table*}[t!]
\centering
\caption{Comparison of our method with existing approaches on video question answering benchmarks. EFlow delivers strong performance across diverse video understanding tasks, with particularly pronounced gains on long-video benchmarks.}
\label{tab:long_video_results}

\resizebox{0.96\textwidth}{!}{
\begin{tabular}{l ccccc}
\toprule
\multirow{2}{*}{\textbf{Methods}} & \multicolumn{2}{c}{\textbf{VideoMME}} & \textbf{LVBench} & \textbf{LongVideoBench} & \textbf{NextGQA} \\
\cmidrule(lr){2-3}
 & Long & Overall & Avg & Avg & Overall \\
\midrule
\multicolumn{6}{l}{\textit{Open-source Single-Turn Video MLLMs}} \\
\midrule
Qwen2.5-VL~\cite{qwen25vl} & 51.0 & 62.9 & 45.3 & 56.0 & 59.5 \\
LLaVA-Video~\cite{llavavideo} & 50.6 & 62.6 & 41.5 & 58.2 & - \\
LLaVA-OneVision~\cite{llavaonevision} & 46.7 & 56.2 & - & 56.4 & - \\
Video-R1~\cite{videor1} & - & 61.4 & - & - & - \\
Rewatch-R1~\cite{rewatchr1} & - & 65.6 & 43.3 & - & - \\
Video-Thinker~\cite{wang2025video} & - & - & 37.0 & - & - \\
VideoChat-R1~\cite{videochatr1} & - & - & - & - & 70.6 \\
Qwen3-VL*~\cite{qwen3vl} & 58.6 & 67.9 & 50.6 & 59.1 & 79.1 \\

\midrule
\multicolumn{6}{l}{\textit{Open-source Agent-based Framework}} \\
\midrule
VideoAgent~\cite{videoagent} & - & - & 29.3 & - & - \\
VideoTree~\cite{videotree} & 54.2 & - & 28.8 & - & - \\
\midrule
\multicolumn{6}{l}{\textit{Open-source Native Multi-turn Tool Invocation Video MLLMs}} \\
\midrule
Conan~\cite{conan} & - & 60.5 & 39.2 & 56.6 & - \\
LongVT~\cite{longvt} & - & 64.3 & 41.3 & - & - \\
Video-Zoomer~\cite{videozoomer} & - & 65.2 & 41.5 & 57.7 & - \\
VITAL~\cite{vital} & 54.0 & 64.1 & - & - & 78.7 \\
\midrule
\textbf{EFlow} &  \textbf{60.1} & \textbf{69.1} & \textbf{52.5} & \textbf{60.3} & \textbf{80.0} \\
\bottomrule
\end{tabular}
}
\end{table*}

\section{Experiments}\label{sec:experiments}

To evaluate the effectiveness of EFlow, we conduct extensive experiments on multiple video understanding benchmarks spanning various video lengths. We aim to answer the following research questions: (1) Does explicit evidence flow through T-CoT and R-CoT improve overall performance? (2) How effective is the margin-based reflection mechanism in handling complex video queries?

\subsection{Experimental Setup}\label{sec:exp:setup}

\paragraph{Implementation Details.} We initialize EFlow from the Qwen3-VL-8B-Instruct checkpoint and perform continued post-training. During the SFT stage, we train the model for 1 epoch with a learning rate of $2e-5$ using a cosine learning rate scheduler. For the RL stage, we employ Group Relative Policy Optimization (GRPO) with a learning rate of $1e-6$ and a group size of $G=16$. According to the dynamic confidence heuristic, the confidence threshold $\tau$ for triggering reflection is empirically adjusted depending on the benchmark complexity. In our ablation studies, we experiment with representative thresholds (e.g., $0.2$, $0.3$, $0.5$) to observe the system's behavior (meaning reflection is triggered if the Confidence Score $c < \tau$), ensuring it activates when the model's leading answer choice is not significantly more probable than its runner-up. During inference, the coarse video representation $\mathcal{V}_{coarse}$ is sampled at 1 frame per second (FPS), while the cropped high-resolution clips $\mathcal{V}_{clips}$ are sampled at 1 FPS to capture fine-grained details.


\subsection{Main Results}\label{sec:exp:main}

Table~\ref{tab:long_video_results} presents the quantitative comparison between EFlow and strong video MLLMs across various benchmarks.

\paragraph{Long Video Understanding.}
EFlow achieves competitive performance across all evaluated benchmarks. On VideoMME (Overall, w/o subtitle), which covers videos of varying lengths, our model achieves 69.1\%, outperforming strong multi-turn baselines like Video-Zoomer (65.2\%) and VITAL (64.1\%). The advantage is especially pronounced on long-video benchmarks: EFlow reaches 60.1\% on VideoMME (Long), 52.5\% on LVBench, and 60.3\% on LongVideoBench, substantially surpassing all baselines. On the temporal grounding benchmark NextGQA, EFlow also achieves 80.0\%, confirming the generality of the staged flow design. These results demonstrate that combining GRPO reinforcement learning with explicit evidence flow yields a robust video agent that excels across the full spectrum of video lengths.

\paragraph{Complex Video Reasoning.}
To demonstrate that explicit evidence flow also enhances reasoning capabilities without requiring extensive data fine-tuning, and offers even greater value as base model capabilities improve, we conduct an exploratory experiment on complex video reasoning tasks. As shown in Table~\ref{tab:reasoning}, EFlow demonstrates consistent improvements over baseline models across VSI-Bench~\cite{vsibench} and Video-MMMU~\cite{videommmu}. This validates that routing temporal grounding and reasoning through connected flow stages prevents error cascading, enabling the model to focus its reasoning capacity on the correctly grounded visual evidence.

\begin{table}[t]
\centering
\caption{Comparison with representative methods on video reasoning benchmarks.}
\label{tab:reasoning}
\resizebox{\columnwidth}{!}{
\begin{tabular}{lcc}
\toprule
\textbf{Model} & \textbf{VSI-Bench} & \textbf{Video-MMMU} \\
& \textbf{Acc} & \textbf{Acc} \\
\midrule
LongVA-7B~\cite{longva} & 29.2 & 23.9 \\
VILA-1.5-8B~\cite{vila} & 28.9 & 20.8 \\
LLaVA-OV-7B & 32.4 & 33.8 \\
AoTD-7B~\cite{aotd} & 28.8 & — \\
Video-R1-7B ~\cite{videor1} & 37.1 & 52.4 \\
VideoRFT-7B~\cite{videorft} & 36.8 & 51.1 \\
Temporal-RLT-7B~\cite{temporalrlt} & — & — \\
Qwen2.5-VL-7B\cite{qwen25vl} & 31.8 & 47.4 \\
 VITAL-7B\cite{vital} & 41.8 & 54.2 \\
 Qwen3-VL-8B\cite{qwen3vl} & 57.9 & 65.3 \\
 \midrule
 \textbf{EFlow} & \textbf{59.1} & \textbf{65.5} \\
 \midrule
\bottomrule
\end{tabular}
}
\end{table}

\subsection{Ablation Studies}\label{sec:exp:ablation}

To understand the contribution of each component in EFlow, we conduct a series of ablation studies on representative benchmarks.

\paragraph{Effect of Multi-Stage Training.}

We first study the contribution of each training stage. Starting from the vanilla Qwen3-VL (8B) backbone, we progressively add Supervised Fine-Tuning (SFT), Reinforcement Learning (RL), and Reinforcement Fine-Tuning (RFT). As shown in Table~\ref{tab:ablation_training}, SFT alone already yields a clear improvement over the base model by teaching the agent how to follow the T-CoT / R-CoT template and correctly invoke tools. While RL alone does not consistently improve over SFT on all metrics, it refines the model's exploration policy for temporal grounding and reflection decisions, laying the groundwork for the next stage. Finally, RFT distills high-reward trajectories from RL back into supervised data, consolidating the learned behaviors and delivering a substantial and stable gain across all datasets.

\paragraph{Effect of Adaptive Reflection.}

\begin{figure}[!ht]
\centering
\includegraphics[width=0.8\columnwidth]{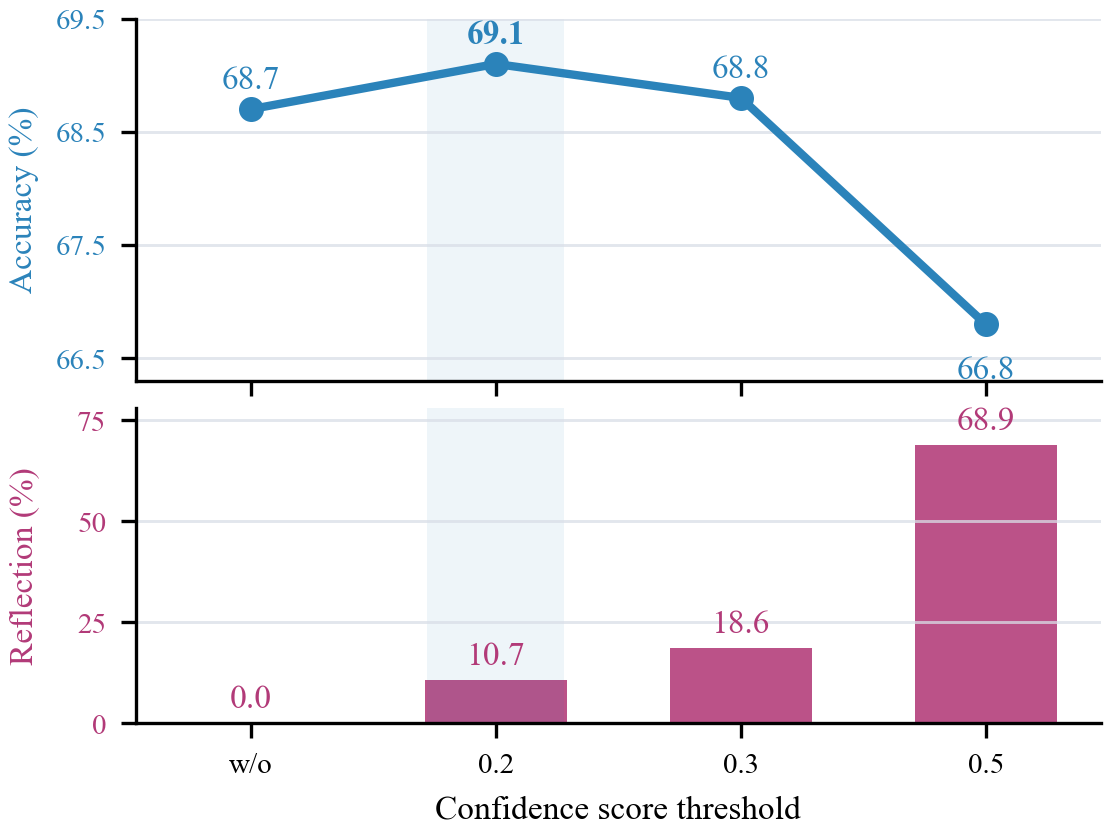}
\caption{Effect of margin-based reflection. A moderate confidence threshold improves accuracy while keeping the reflection rate low; an overly loose threshold triggers reflection too frequently and degrades performance.}
\label{fig:ablation_reflection}
\end{figure}

Finally, we analyze the impact of the competition-based reflection mechanism by varying the confidence threshold $\tau$ (see Figure~\ref{fig:ablation_reflection}). Without reflection, EFlow performs competitively but suffers from occasional error cascading if initial crops miss critical evidence. An optimal tight threshold ($\tau = 0.2$) achieves the best accuracy. It triggers reflection very selectively (10.7\%), successfully instructing the model to strategically ``re-read'' the global video only when genuinely hesitant. As the tolerance increases slightly ($\tau = 0.3$), reflection triggers more frequently (18.6\%), leading to sub-optimal but reasonable performance. Conversely, when the threshold becomes too loose ($\tau = 0.5$), excessively more samples trigger reflection (68.9\%). This forces the model to frequently fall back to the coarse video, discarding fine-grained clip details and resulting in the worst performance. Given these dynamics, our framework adopts a dynamic thresholding approach rather than a rigid value to optimally balance accuracy and efficiency across diverse testing scenarios. This pattern also clarifies the role of reflection in our framework: it is not intended to replace local evidence with global context, but to act as a sparse repair mechanism when the local reasoning path is unreliable. By keeping reflection selective, EFlow preserves the benefit of focused clips while still retaining a route for recovery from poor temporal grounding. This explains why the moderate threshold improves both robustness and efficiency, whereas excessive reflection weakens the very evidence concentration that temporal cropping is designed to provide.

\begin{table}[t!]
\centering
\caption{Effect of multi-stage training on video understanding. Accuracies (\%) are reported.}
\label{tab:ablation_training}

\resizebox{\columnwidth}{!}{
\begin{tabular}{lcc}
\toprule
\textbf{Model} & \textbf{VideoMME (Long)} & \textbf{LVBench (Avg)} \\
\midrule
Qwen3-VL (8B) & 58.6 & 50.6 \\
EFlow-SFT & 59.0 & 51.2 \\
EFlow-RL & 59.6 & 51.2 \\
\textbf{EFlow (Ours)} & \textbf{60.1} & \textbf{52.5} \\
\bottomrule
\end{tabular}
}

\end{table}

\section{Conclusion}\label{sec:conclusion}

This paper presents EFlow, a tool-augmented framework that mitigates error cascading in video understanding by learning and executing an explicit evidence flow. A competition-margin guided reflection mechanism enables confidence-based adaptive fallback to global context. Experiments on five benchmarks demonstrate strong performance across diverse video lengths, with particularly strong gains on long videos. 

\clearpage
\section{Limitations}\label{sec:limitations}

EFlow relies on the backbone model's ability to localize relevant visual evidence from long videos. If the underlying model cannot identify plausible temporal regions, explicit evidence flow may still propagate imperfect grounding. For this reason, our main evaluation is conducted on Qwen3-VL, a strong video MLLM with capable temporal localization, and generalization to weaker or substantially different backbones remains an important direction for future study. In addition, Adaptive Reflection depends on a useful uncertainty prior from the answer confidence signal. When this signal is poorly calibrated, reflection may require task-specific priors, threshold tuning, or multiple validation passes to decide whether re-reading the global video is truly beneficial.



\bibliography{main}

\clearpage

\end{document}